\documentclass[runningheads]{llncs}

\usepackage{eccv}

\usepackage{eccvabbrv}

\usepackage{graphicx}
\usepackage{booktabs}

\usepackage[accsupp]{axessibility}  

\usepackage{hyperref}

\usepackage{orcidlink}

\begin{document}

\title{CosmosAlign: Adapting a World Foundation Model for Generative Traffic Video Forecasting} 

\titlerunning{CosmosAlign for Generative Traffic Video Forecasting}

\author{Quang Minh Dinh\inst{1}\orcidlink{0009-0003-9205-6270} \and
Tuan Kiet Doan\inst{2}\orcidlink{0009-0009-2717-0257}}

\authorrunning{Q. M. Dinh and T. K. Doan}

\institute{Simon Fraser University, Burnaby, BC, Canada\\
\email{qmd@sfu.ca} \and
Institut Polytechnique de Paris, Palaiseau, France\\
\email{tuan.doan@ip-paris.fr}}

\maketitle

\begin{abstract}
Generative traffic video forecasting aims to synthesize long-horizon, temporally coherent future videos of traffic scenes from a short observation history and textual descriptions. In this paper, we present CosmosAlign, a generative traffic video forecasting framework built upon the pretrained Cosmos3-Nano world foundation model. Our approach is motivated by the observation that successfully adapting large pretrained world models to downstream forecasting tasks depends primarily on distribution alignment rather than increased model capacity. To this end, we propose a two-stage LoRA adaptation strategy that first aligns the conditioning-mode distribution with the target forecasting task, and then aligns the training captions with the model's native structured prompting interface through an LLM-based re-captioning pipeline. During inference, we further improve prediction quality using a fully training-free procedure consisting of consensus-based medoid sample selection and motion-adaptive blending of static scene regions. CosmosAlign achieves a final score of 76.49 on the AI City Challenge 2026 Track 5 benchmark, ranking first on the final leaderboard. Our code is publicly available at \url{https://quangminhdinh.github.io/CosmosAlign/}

\keywords{Traffic Video Forecasting \and World Models \and Autonomous
Driving \and Parameter-Efficient Fine-Tuning}
\end{abstract}

\section{Introduction}
\label{sec:intro}
The goal of generative video forecasting (GVF) is to generate future video frames from a
sequence of observed context frames~\cite{oprea2020review}, optionally guided by control
signals such as text prompts or camera motion~\cite{Ho2022VideoDiffusion, ming2024survey}.
A key advantage of GVF lies in its ability to capture complex spatio-temporal dynamics,
making it an essential step toward building internal world models. Thanks to this
predictive capability, GVF has been successfully applied across a wide range of
applications such as robotic control~\cite{ebert2018visual}, planning~\cite{xie2019improvisation},
and anomaly detection~\cite{liu2018future}.

In autonomous driving, GVF has evolved beyond simple frame prediction into driving world
models~\cite{zhou2026drivinggen, hu2309gaia}. Given an initial observation and conditioning
signals, these models predict not only the future
movement of the ego vehicle but also the dynamic behavior of the surrounding traffic
participants. By supporting simulation and generating diverse synthetic data,
they reduce the dependence on real-world data collection and provide a safe way to test
rare scenarios~\cite{hassan2025gem, mousakhan2026overcoming, gao2024vista}.

However, since these simulators feed end-to-end autonomous driving systems, inaccuracies
in the predicted scenes and trajectories can propagate directly into real-world driving
safety~\cite{wang2023efficient, shao2023safety, shao2024lmdrive}. Improving the physical
plausibility, long-range temporal consistency, and lifelike behavior of generated traffic
scenarios therefore remains a critical problem. In response to this challenge, the AI City Challenge 2026~\cite{Tang26AICity26} establishes a benchmark for text-conditioned traffic video forecasting. Given a short
history of observed frames and textual descriptions of the scene, participants are
required to generate the future frames of pedestrian-vehicle interaction scenarios. Built
on the Woven Traffic Safety (WTS) dataset~\cite{Kong2024WTS}, this setting directly tests
whether generative models can synthesize futures that are not only visually faithful
but also temporally coherent and consistent with the described behavior in safety-critical
situations.

Recent large-scale foundation world models, such as the GAIA~\cite{hu2309gaia,
russell2025gaia}, DriveDreamer~\cite{wang2024drivedreamer, zhao2025drivedreamer},
DrivingWorld~\cite{hu2024drivingworld}, Genie~\cite{bruce2024genie}, and Cosmos~\cite{agarwal2026cosmos, ali2025world} families, offer a promising foundation for traffic video forecasting.
By scaling model capacity and training data, they simulate complex physical dynamics and
maintain temporal coherence under text conditioning. Yet our study reveals that adapting
such generalist world models to a specific forecasting benchmark is less a question of
capacity than of alignment. To address this, we present CosmosAlign, a framework built on the Cosmos3-Nano world model. CosmosAlign aligns both the task conditioning distribution and prompt formatting using a two-stage parameter-efficient adaptation strategy, complemented by a training-free inference procedure for robust frame generation. Our contributions are summarized as follows:
\begin{itemize}
    \item We identify distribution alignment, rather than model capacity, as the primary challenge in adapting large pretrained world models to traffic video forecasting. Based on this insight, we propose a two-stage LoRA adaptation strategy that aligns both the conditioning-mode distribution and the prompt representation with the target forecasting task.

    \item We introduce an inference pipeline consisting of consensus-based sample selection and motion-adaptive blending, improving robustness and visual fidelity without requiring auxiliary models or ground-truth.

    \item We conduct extensive ablation studies across fine-tuning, prompting, sampling, and post-processing, demonstrating that alignment-oriented design choices consistently yield larger gains than increasing model capacity or modifying the generation process.

    \item Our approach achieves a final score of 76.49 on the AI City Challenge 2026 Track 5 benchmark, ranking first on the final leaderboard while achieving the best PSNR and LPIPS and the joint-highest SSIM.
\end{itemize}

\section{Related Work}

\textbf{World Models for Autonomous Driving. }A world model~\cite{ha2018world} is an internal
simulator of environmental dynamics that supports predictive and counterfactual rollouts for
sensory understanding, trajectory forecasting, and autonomous decision-making~\cite{li2025comprehensive}.
Two main research
directions have emerged~\cite{ali2025world}. On the one hand, predictive models in latent
representation spaces~\cite{hafner2019dream, hafner2019learning}, including JEPA-style
architectures~\cite{lecun2022path, bardes2024revisiting, assran2025v}, compress
high-dimensional inputs into compact latents that align well with perception, prediction,
and planning while offering high efficiency. On the other hand, pixel-space world
models frame environment simulation as a generative video modeling
task~\cite{bruce2024genie, agarwal2025cosmos, agarwal2026cosmos}. By preserving
high-fidelity detail, such models double as synthetic data generators for downstream policies. 

In autonomous
driving, world models must further satisfy strict requirements, including the complex dynamics
of traffic participants, ego-motion control, and cross-view consistency~\cite{russell2025gaia}.
Early works such as GAIA-1~\cite{hu2309gaia} and CommaVQ~\cite{commaai2023commavq} are often
restricted to a single camera view and limited text- or action-based control, while later
works like Drive-WM~\cite{wang2024driving} and UniMLVG~\cite{chen2025unimlvg} enable multi-view
generation conditioned on a wide range of inputs, and MaskGWM~\cite{ni2025maskgwm} and
Vista~\cite{gao2024vista} focus on long-duration, high-resolution generation. Beyond 2D video,
DriveDreamer4D~\cite{zhao2025drivedreamer} and DreamDrive~\cite{mao2025dreamdrive} combine
generative models with real-world videos to construct interactive 4D environments for
closed-loop testing. More recently, general-purpose physical AI foundation models such as
Cosmos~3~\cite{agarwal2026cosmos} and Genie~\cite{bruce2024genie} have shown remarkable
effectiveness on driving tasks, motivating our choice of adapting such a model in this work.

\textbf{Multimodal Learning for Driving. }Together with the advancements of
vision-language pre-training~\cite{radford2021clip, li2023blip2, liu2023llava}, recent works
are moving toward adapting multimodal foundation models to the driving domain, grounding
visual observations in natural language to gain the interpretability that traditional
perception-control pipelines lack~\cite{wen2023gpt4v}. Early efforts repurpose general
vision-language models for driving scene understanding, visual question answering, and
captioning~\cite{wen2023gpt4v, ma2024dolphins}, while later works move toward interpretable
decision-making, such as DriveGPT4~\cite{xu2024drivegpt4}, which jointly predicts control
signals and language explanations, and DriveLM~\cite{sima2024drivelm}, which casts driving
reasoning as graph visual question answering. Closest to our setting, a line of research focuses on fine-grained traffic-safety analysis in critical pedestrian-vehicle interactions. For instance, TrafficVLM~\cite{dinh2024trafficvlm} reformulates safety modeling as joint temporal localization and dense captioning. CityLLaVA~\cite{duan2024cityllava} and
TrafficInternVL~\cite{wu2025trafficinternvl} further refine visual prompting and structured
fine-tuning on the WTS benchmark.


\textbf{Foundation Model Fine-Tuning. }While large-scale foundation models generalize
remarkably well, fully fine-tuning them under the strict hardware budgets of autonomous driving
is computationally impractical and risks catastrophic forgetting of the pretrained physical
priors. To reduce the memory footprint of fine-tuning, a range of methods have been proposed,
spanning activation compression~\cite{miles2024velora, nguyen2025beyond}, optimizer
compression~\cite{zhao2024galore, muhamed2024grass}, and parameter-efficient fine-tuning
(PEFT)~\cite{zhou2024autopeft, wang2022adamix}. Among them, LoRA and its
variants~\cite{Hu2022LoRA, liu2024dora, meng2024pissa} have become the most widely adopted: by
freezing the pretrained weights and training only two additional low-rank matrices, they strike
a favorable balance between computational efficiency and model capability. Given its robust
empirical success, we adopt LoRA to efficiently adapt the foundation model in this work.


\section{Method}
\label{sec:method}

Generative Traffic Video Forecasting is a challenging task which involves synthesizing a long
future continuation of a traffic safety scenario, given only a short clip of observed history
frames and textual descriptions of the scene. The generated frames must remain faithful to
the observed scene layout, exhibit realistic pedestrian and vehicle motion, and stay on the
ground-truth timeline for up to 120 frames. In this section, we present our solution, which
adapts the Cosmos3-Nano world model to this task with a two-stage LoRA
fine-tuning procedure and a carefully aligned inference pipeline, followed by a test-time
sample selection and blending step. In Sec.~\ref{sec:finetuning}, we describe the two fine-tuning stages. We detail the
construction of our structured prompts and negative prompts in Sec.~\ref{sec:prompts}, and
present the full inference and test-time procedure in Sec.~\ref{sec:inference}.

\subsubsection{Problem Formulation.}
Given a history clip $H=\{x_{-n},\dots,x_{0}\}$ of $n$ observed frames at resolution $1280\times720$,
a pair of captions $(c^{p}, c^{v})$ describing the pedestrian and the vehicle, and a target
horizon $N \in [51, 120]$, the goal is to generate $\hat{Y}=\{\hat{x}_{1},\dots,\hat{x}_{N}\}$,
the next $N$ frames of the same scene at the same resolution and frame rate. We adapt the task to the pre-training and mid-training paradigms of Cosmos3-Nano by limiting the history clip $H$ to the last 5 frames.

\begin{figure}[t]
\centering
\includegraphics[width=\linewidth]{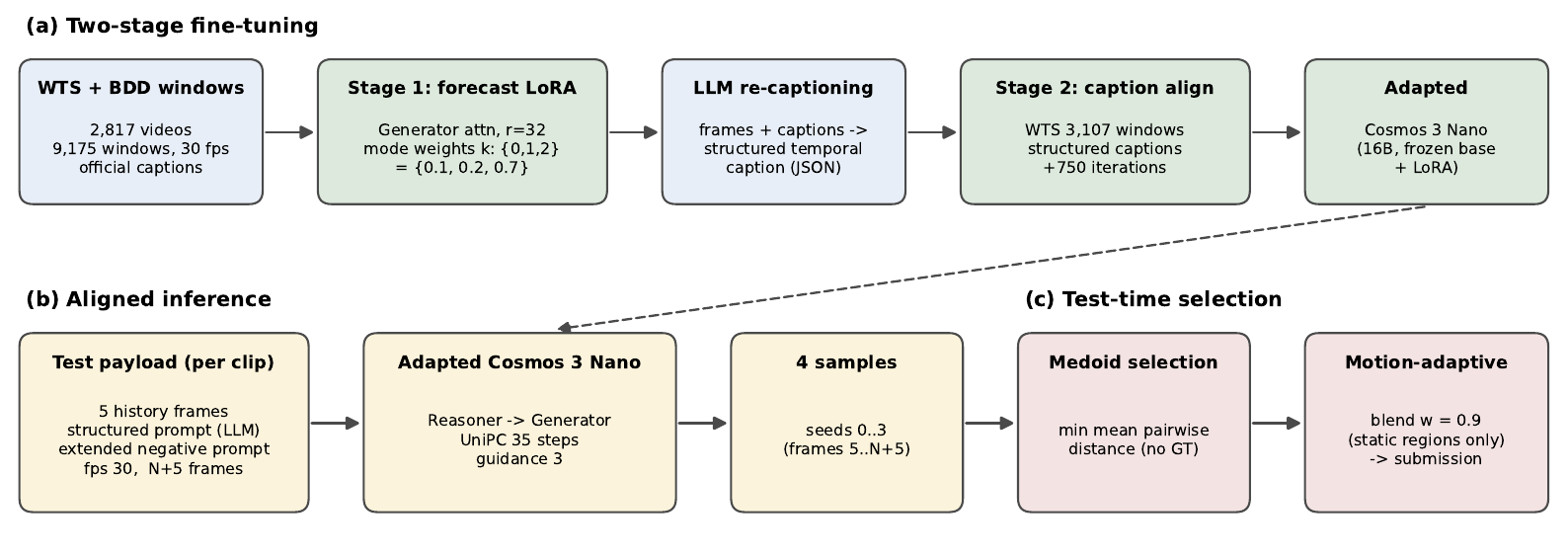}
\caption{Overview of our method. (a) Cosmos3-Nano is adapted in two LoRA stages: Stage 1
fine-tunes the Generator tower on WTS and BDD forecasting windows with the reweighted conditioning-mode
distribution, and Stage 2 continues
training on the WTS windows only, using the structured
temporal-caption format the model natively expects. (b) At test time, each clip is encoded
into a payload holding the five history frames, an LLM-generated structured prompt, an
extended negative prompt, and the native frame rate, from which the adapted model generates
the full horizon in a single pass; four samples are drawn with different seeds. (c) The final
prediction is the medoid of the four samples, blended toward the last observed frame in
static regions only.}
\label{fig:pipeline}
\end{figure}

\subsection{Cosmos 3 World Foundation Models}
\label{sec:basemodel}

Cosmos 3 is a family of omnimodal world foundation models built on a
Mixture-of-Transformers architecture with two coupled towers: a \emph{Reasoner}, an
autoregressive vision-language model that ingests the text prompt and the conditioning
frames, and a \emph{Generator}, a diffusion transformer that denoises future frames
conditioned on the Reasoner's latents. We use the 16B Cosmos3-Nano variant, whose native resolution
720p at scale 16:9 matches the WTS resolution exactly. Video is processed by the VAE encoder from Wan2.2-TI2V5B~\cite{Wang2025WanOA} with 4 times temporal compression, and generation is trained with a
rectified-flow objective: the first $k$ latent frames of a clip are kept
clean as conditioning while the remaining frames are noised, and the velocity-prediction
loss is applied to the non-conditioning frames only. The conditioning length $k$ is drawn
per training sample from a categorical distribution over $k \in \{0,1,2\}$, where $k=0$ implies that the model infer the frames purely from text, and $k=2$
corresponds to the first five conditional pixel frames. Sampling uses a
35-step UniPC solver~\cite{zhao2023unipc} with classifier-free guidance. The model is
frame-rate aware: the \texttt{fps} field of the inference payload conditions generation
through the prompt template and the temporal position encodings.

\subsection{Two-Stage Fine-Tuning}
\label{sec:finetuning}

\subsubsection{Stage 1: Forecasting Adaptation.}
We convert the WTS training pool and the provided BDD external pool into 2{,}817 videos (635
multi-view WTS videos and 2{,}182 BDD dashcam videos), segmented by the annotated scenario
phases into 9{,}175 training windows. Windows keep the native frame rate, and each window
carries the concatenated pedestrian and vehicle captions of its clip as the text prompt. We capped the packed sequences at 10{,}240
tokens, which fits training of the 16B model on a single A100-80GB GPU with full activation
checkpointing. The conditioning-mode distribution of $\{k=0,\; k=1,\; k=2\}$ are reweighted from the default
$\{0.7,\; 0.2,\; 0.1\}$ to
$\{0.1,\; 0.2,\; 0.7\}$, so that 70\% of the gradient steps
supervise the five-frame video-conditioned forecasting mode that the track
evaluates, while a small amount of text-only and single-image conditioning is retained. We
train LoRA adapters of rank 32 ($\alpha=64$) on the query, key, value, and
output projections of the Generator tower attention layers, keeping the base model and the
Reasoner frozen.

\subsubsection{Stage 2: Structured Caption Alignment.}
Cosmos3-Nano is post-trained to expect structured prompts, a JSON payload with a multi-sentence
temporal caption narrative plus duration, frame-rate, and resolution fields,
rather than free-form caption strings. To align the adapter with this interface, we
re-caption all 3{,}107 WTS training windows with the LLM pipeline in Sec.~\ref{sec:prompts}
and serialize each caption identically to the inference payload format. We continue training on this WTS-only structured corpus starting from the Stage 1 checkpoint.

\subsection{Prompt Construction}
\label{sec:prompts}

\begin{figure}[t]
\centering
\fbox{\begin{minipage}{0.94\linewidth}\scriptsize
You caption traffic-safety video clips for a video generation model. In each request you
receive: the camera type (overhead surveillance camera or vehicle-mounted dashboard
camera), the clip duration in seconds, the official pedestrian and vehicle captions of the
scenario, and frames sampled from the clip in temporal order.\\[3pt]
Write one temporal caption of five to eight sentences, in present tense, that narrates the
clip from its first frame to its last:
\begin{enumerate}\setlength{\itemsep}{1pt}
\item Open by establishing the viewpoint and the scene. For overhead cameras, describe the
scene as seen from above; for vehicle cameras, open with ``From inside a vehicle, \dots''
or ``From a forward-facing dashboard camera mounted in a moving car, \dots''.
\item Introduce the visible agents with their appearance and positions (clothing,
approximate age, vehicle color and type), exactly as visible in the frames.
\item Describe how the scene evolves in temporal order (``As time passes, \dots''). Take
the agents' behavior over time from the official captions; take every visual detail
(lighting, weather, road surface, markings, signage, camera stillness or motion) from the
frames.
\item Ground every statement in the provided frames or captions. Never invent objects,
agents, events, or camera motion that they do not support. If the captions and the frames
disagree, trust the frames.
\item Use no meta-language (``in this video''), do not mention the captions or the frames,
and do not address the viewer.
\end{enumerate}
Return only the caption text, with no preamble and no formatting.
\end{minipage}}
\caption{System prompt of our caption-generation pipeline.}
\label{fig:sysprompt}
\end{figure}

\begin{figure}[t]
\centering
\fbox{\begin{minipage}{0.985\linewidth}\scriptsize\ttfamily\raggedright
\{"temporal\_caption": "A broad asphalt driving course is seen from above, lined with
black-and-yellow cones and edged by grass strips, distant parked cars, and trees under
bright, clear daylight. A man in his twenties in a brown jacket and navy-blue slacks walks
across the open pavement with a dark car standing close behind him, both facing the same
direction. As time passes, the car eases straight forward at a steady speed while the man
continues ahead at a slow, even walk, holding his heading. [\dots]",
"duration": "2.9", "fps": "30", "resolution": \{"H": 720, "W": 1280\},
"aspect\_ratio": "16,9"\}
\end{minipage}}
\caption{Example structured prompt of a test clip (overhead view, horizon
$N=87$).}
\label{fig:promptexample}
\end{figure}

\begin{figure}[t]
\centering
\fbox{\begin{minipage}{0.985\linewidth}\scriptsize
Pedestrians and vehicles that should move remain frozen in place like statues for the
entire duration. Moving figures slowly melt, smear, and morph, limbs dissolving with ghost
duplicates lingering. The color grade drifts steadily toward a warm sepia wash, and halos
around streetlights bloom progressively larger. The fixed surveillance camera drifts and
creeps when it should be perfectly still.
\end{minipage}}
\caption{The four sentences appended to the default Cosmos3-Nano negative prompt.}
\label{fig:negprompt}
\end{figure}

\subsubsection{Structured Prompt Generation.}
We generate all structured prompts for the 3{,}107 Stage 2 training windows and the 71
test clips using Claude Opus 4.8 API, which can be replaced by any Visual Language Model (VLM) of equivalent capability. For each video, the VLM receives the camera type, the clip duration, the official
pedestrian and vehicle captions, and conditional frames sampled in temporal order, taken from the
training window itself for Stage-2 data and the provided history frames
for test clips. The system prompt, shown in Fig.~\ref{fig:sysprompt}, limits the output
to 5-8 sentences in the \texttt{temporal\_caption} field of the base model's native
captioning format, constrains every visual detail to be grounded in the frames and every
behavioral statement in the official captions, and forbids inventing unsupported content.
The returned caption is serialized together with the clip's duration, frame-rate,
resolution, and aspect-ratio fields (Fig.~\ref{fig:promptexample}), matching the
training-side serialization.

\subsubsection{Negative Prompt.}
We keep the full default structured negative prompt that ships with Cosmos3-Nano, which lists
generic generation defects (blurry subjects, compression artifacts, broken physics), and
append the four sentences shown in Fig.~\ref{fig:negprompt}, which target the failure
modes we observed in generated traffic video: frozen agents, melting and morphing figures,
drifting color grade, blooming light halos, and camera creep on fixed cameras. The same
extended negative prompt is used for every clip in both views.

\subsection{Inference and Test-Time Procedure}
\label{sec:inference}

\subsubsection{Generation.}
Each test clip is encoded into a payload holding the last five conditional history frames, the structured prompt, the extended negative prompt,
$\texttt{num\_frames} = N+5$, and $\texttt{fps} = 30$, the native frame rate of both the
fine-tuning windows and the provided test histories. The adapted model generates the whole
clip in a single pass with the 35-step UniPC solver and guidance scale $g=3$.
We draw four samples per clip with seeds 0 to 3.

\subsubsection{Medoid Sample Selection.}
Inspired by Minimum Bayes Risk decoding in text generation~\cite{kumar-byrne-2004-minimum,eikema-aziz-2020-map} and self-consistency in language-model reasoning~\cite{Wang2022SelfConsistencyIC}, we employ a consensus-based selection rule.
From the four samples $\{\hat{Y}^{(1)},\dots,\hat{Y}^{(4)}\}$ of a clip, we select the
medoid, the sample with the minimum mean distance to the other samples:
\begin{equation}
  \hat{Y}^{*} = \operatorname*{arg\,min}_{i}\; \frac{1}{3}\sum_{j \neq i}
  d\!\left(\hat{Y}^{(i)}, \hat{Y}^{(j)}\right),
\label{eq:medoid}
\end{equation}
where $d$ is the mean absolute difference between temporally aligned frames on a
spatially downsampled grid. The selection uses no ground-truth information

\subsubsection{Motion-Adaptive Blending.}
Finally, each selected prediction is blended toward the last conditional frame $x_0$ in the
regions where the generation itself is static. We compute a per-pixel motion map
\begin{equation}
  m(u) = \frac{1}{N}\sum_{t=1}^{N} \left| \bar{\hat{x}}_t(u) - \bar{x}_0(u) \right|,
\end{equation} 
where
$\bar{x}$ denotes grayscale intensity, and map it to a blending weight
\begin{equation}
  w(u) = 0.9 \cdot \operatorname{clip}\!\left(\frac{m_{hi} - m(u)}{m_{hi} - m_{lo}},\, 0,\, 1\right),
  \qquad m_{lo}=3,\; m_{hi}=20,
\label{eq:blend}
\end{equation}
smoothed with a Gaussian kernel ($\sigma=8$) to avoid seams. The final frames are:
\begin{equation}
  \hat{x}_t'(u) = \bigl(1 - w(u)\bigr)\,\hat{x}_t(u) + w(u)\,x_0(u)
\end{equation}
The weight is high only
where the generated clip stays close to the last conditional frame for its whole duration, so
moving agents are left untouched, and on dashcam clips with ego-motion the weight vanishes
everywhere. The blending is therefore self-regulating across the two camera types.

\subsubsection{Temporal Deflickering.}
As a lighter alternative to the blending step, we also evaluate a motion-compensated
temporal low-pass against flicker, the high-frequency temporal noise in which fine
textures and colors change slightly from frame to frame even where the scene is static.
We estimate the optical flow
between frame $t$ and each of its neighbors with RAFT~\cite{Teed2020RAFTRA} and warp the neighbors
onto frame $t$, so that corresponding pixels coincide before averaging. Each interior
frame is then replaced by
\begin{equation}
  \hat{x}_t' = (1 - 2\beta)\,\hat{x}_t
  + \beta\,\mathcal{W}_{t-1 \to t}(\hat{x}_{t-1})
  + \beta\,\mathcal{W}_{t+1 \to t}(\hat{x}_{t+1}),
  \qquad \beta = 0.2,
\label{eq:deflicker}
\end{equation}
where $\mathcal{W}_{s \to t}$ denotes warping frame $s$ onto frame $t$ along the estimated
flow. Static regions are thus averaged over three aligned observations of the same
content, which cancels the temporal noise, while moving content is realigned before
averaging and stays sharp. Stacked on best-of-3 selection, this filter produced our
second-best submission. Our final submission uses the motion-adaptive blend instead, and
we did not combine the two.

\section{Experiments}
\label{sec:experiments}

\subsection{Experimental Setup}
\label{sec:setup}

\subsubsection{Datasets.}
We use a subset of the WTS dataset provided by the Track 5 of the AI City Challenge 2026, which contains staged
traffic safety scenarios recorded simultaneously from fixed overhead cameras and vehicle
cameras at resolution $1280\times720$, along with external vehicle camera videos extracted from the BDD100K
dataset~\cite{Yu_2020_CVPR}. Our Stage 1 corpus holds 2{,}817
videos (635 WTS, 2{,}182 BDD) segmented into 9{,}175 phase-aligned windows at the native
frame rate. The Stage 2 corpus is the WTS subset (3{,}107 windows) with re-generated
structured captions. The test set contains 71 clips with 10 to 224 provided history
frames each, requested horizons of 51 to 120 frames (mean 73), and roughly 38\% overhead
views. For local model selection we hold out a validation set of 36 WTS clips (14 overhead,
22 vehicle, matching the test view ratio), built to mirror the test protocol exactly:
histories of five consecutive native-rate frames and per-clip horizons drawn from the
empirical test horizon distribution, with payloads identical to test payloads in every field
except the one under study.

\subsubsection{Evaluation Metrics.}
Following the evaluation setup of Track~5, we adopt PSNR, SSIM, LPIPS~\cite{Zhang2018LPIPS},
CLIP score~\cite{Hessel2021CLIPScoreAR}, FID~\cite{Heusel2017GANsTB}, and FVD~\cite{Unterthiner2018FVD} as the evaluation metrics,
normalized and averaged with equal weights by the AI City Challenge's evaluation system into a single final
score. Our local harness computes PSNR, SSIM, LPIPS, CLIP-S, and FVD over the 36 validation
clips. The local CLIP-S is on a 0 to 100 scale while the server reports 0 to 1.

\subsubsection{Implementation Details.}
All fine-tuning and inference run on a single NVIDIA A100-80GB GPU. Stage 1 runs for 2{,}000
iterations in about 14 hours with AdamW (learning rate $5{\times}10^{-4}$, cosine
decay, 100 warmup steps, gradient accumulation 4) in bfloat16, full activation
checkpointing. Stage 2 adds up to 1{,}000 iterations in about 8
hours, using the same
optimizer settings and an extended cosine schedule. Inference uses the framework's native pipeline, which shares the training conditioning path, with 35-step UniPC sampling and guidance 3. A single 71-clip test sweep takes about 8 hours per seed, and all configuration
sweeps are validated at seed 0. The \texttt{fps} field of every payload is set to 30 to
match the native frame rate of the fine-tuning windows and of the provided test histories.

\subsection{Ablation Studies}
\label{sec:ablations}

\begin{table}[t]
\centering
\caption{Ablation studies on our validation set. The capacity and post-processing blocks
use plain prompts at default guidance, and the remaining blocks use the full single-seed
inference recipe. LoRA $\alpha$ is twice the rank. ``MA'' denotes
motion-adaptive. In the Stage 1
block, the fidelity metrics differ only at noise level and are not highlighted.}
\label{tab:valablations}
\setlength{\tabcolsep}{2pt}\scriptsize
\begin{minipage}[t]{0.47\linewidth}
\centering
\begin{tabular}{@{}lccccc@{}}
\toprule
Config & PSNR & SSIM & LPIPS & CLIP-S & FVD \\
\midrule

\multicolumn{6}{@{}l}{\textit{History}} \\
5 frames  & \textbf{24.40} & \textbf{0.793} & \textbf{0.163} & \textbf{28.12} & \textbf{17.64} \\
9 frames  & 23.76 & 0.777 & 0.177 & 28.09 & 19.73 \\
13 frames & 23.25 & 0.775 & 0.184 & 28.08 & 20.69 \\
\midrule
\multicolumn{6}{@{}l}{\textit{Stage 1}} \\
2{,}000 iter   & 23.92 & 0.791 & 0.173 & \textbf{27.81} & 17.76 \\
3{,}000 iter   & 23.92 & 0.793 & 0.171 & 27.72 & \textbf{17.15} \\
4{,}000 iter   & 23.92 & 0.793 & 0.170 & 27.58 & 17.56 \\
\midrule
\multicolumn{6}{@{}l}{\textit{Stage 2}} \\
Base   & 23.92 & 0.791 & 0.173 & 27.81 & 17.76 \\
 +250   & 24.19 & 0.796 & 0.167 & 27.81 & 16.94 \\
 +500   & \textbf{24.27} & \textbf{0.798} & \textbf{0.166} & 27.87 & 16.31 \\
 +750   & 24.25 & \textbf{0.798} & \textbf{0.166} & \textbf{27.92} & \textbf{15.98} \\
 +1{,}000 & 24.25 & \textbf{0.798} & \textbf{0.166} & 27.90 & 16.19 \\
\bottomrule
\end{tabular}
\end{minipage}\hfill
\begin{minipage}[t]{0.51\linewidth}
\centering
\begin{tabular}{@{}lccccc@{}}
\toprule
Config & PSNR & SSIM & LPIPS & CLIP-S & FVD \\
\midrule
\multicolumn{6}{@{}l}{\textit{Guidance}} \\
$g=1$ & 23.52 & 0.780 & 0.184 & 27.71 & 19.98 \\
$g=2$ & \textbf{23.94} & \textbf{0.791} & \textbf{0.173} & 27.80 & 18.22 \\
$g=3$ & 23.92 & \textbf{0.791} & \textbf{0.173} & 27.81 & \textbf{17.76} \\
$g=6$ & 23.24 & 0.780 & 0.190 & \textbf{27.83} & 20.48 \\
\midrule

\multicolumn{6}{@{}l}{\textit{Adapter}} \\
rank 32   & \textbf{22.88} & \textbf{0.761} & \textbf{0.193} & 27.53 & \textbf{20.84} \\
rank 128  & 21.81 & 0.716 & 0.214 & \textbf{27.58} & 23.06 \\

\midrule
\multicolumn{6}{@{}l}{\textit{Post-proc.}} \\
none         & 22.88 & 0.761 & 0.193 & 27.53 & 20.84 \\
deflickering & 22.93 & 0.765 & 0.193 & 27.57 & \textbf{20.57} \\
color match  & 23.07 & 0.763 & 0.193 & 27.60 & 21.10 \\
global blend & 23.86 & 0.793 & 0.192 & \textbf{27.67} & 22.71 \\
MA blend  & \textbf{23.94} & \textbf{0.808} & \textbf{0.185} & 27.39 & 23.44 \\
\bottomrule
\end{tabular}
\end{minipage}
\end{table}

\begin{table}[t]
\centering
\caption{Progression of our experiments on the Track~5 test set. ``MA'' denotes motion-adaptive blending. Same indentation denotes different options stack separatedly on top of the previous configuration.}
\label{tab:ladder}
\setlength{\tabcolsep}{4pt}
\begin{tabular}{@{}lccccccc@{}}
\toprule
Configuration & Final$\uparrow$ & PSNR$\uparrow$ & SSIM$\uparrow$ & LPIPS$\downarrow$ & CLIP$\uparrow$ & FID$\downarrow$ & FVD$\downarrow$ \\
\midrule
Stage 1 (rank 32, $\alpha=64$)      & 73.31 & 18.60 & 0.576 & 0.278 & 0.938 & 25.19 & 24.83 \\
+ chunked rollout            & 71.33 & 18.04 & 0.573 & 0.323 & 0.936 & 30.67 & 24.96 \\
+ view-balanced data         & 73.05 & 18.25 & 0.567 & 0.283 & 0.937 & 24.52 & 24.36 \\
+ guidance $g=3$                           & 74.76 & 19.18 & 0.593 & 0.260 & 0.941 & 22.34 & 22.27 \\
\quad+ struct.     & 74.68 & 19.06 & 0.595 & 0.264 & 0.941 & 22.36 & 22.32 \\
\quad\quad+ extended neg.     & 74.76 & 19.09 & 0.596 & 0.263 & 0.942 & 22.28 & 22.02 \\
\quad+ Stage 2            & 75.00 & 19.19 & 0.598 & 0.260 & 0.942 & 21.75 & 21.04 \\
\quad\quad+ deflickering      & 75.03 & 19.26 & 0.603 & 0.260 & 0.941 & 22.19 & 21.09 \\
\quad\quad+ best-of-3         & 75.32 & 19.35 & 0.605 & 0.254 & 0.941 & \textbf{21.51} & \textbf{20.57} \\
\quad\quad\quad+ deflickering      & 75.35 & 19.42 & 0.610 & 0.254 & 0.940 & 21.96 & 20.66 \\
\quad\quad+ best-of-4 + MA   & \textbf{76.49} & \textbf{20.12} & \textbf{0.650} & \textbf{0.246} & \textbf{0.950} & 22.41 & 21.79 \\
\bottomrule
\end{tabular}
\end{table}


\subsubsection{Classifier-free Guidance.}
Tab.~\ref{tab:valablations} sweeps the classifier-free guidance scale on the validation set with
all other settings fixed to the final recipe. The quality curve is U-shaped: $g=1$ washes
out conditioning adherence and degrades every metric, $g=2$ ties $g=3$ on the fidelity
metrics while losing 0.46 FVD, and the framework default $g=6$ loses
on four of five metrics. We select $g=3$; on the test set this contributed $+1.45$ with
all six metrics improving (Tab.~\ref{tab:ladder}).

\subsubsection{Stage 2 Fine-Tuning.}
Tab.~\ref{tab:valablations} evaluates the Stage 2 checkpoints at the full inference recipe. Every
checkpoint beats the Stage 1 adapter on all five metrics. Fidelity plateaus after 500
additional iterations while CLIP-S keeps rising, which is the opposite of the CLIP decay we
observe when simply training Stage 1 longer (see the longer-training ablation below),
indicating adaptation to the prompt format rather than memorization. We select the
checkpoint at iteration 750, which attains the best FVD.

\subsubsection{Best-of-$N$ Selection.}
Tab.~\ref{tab:bestofn} quantifies sampling stochasticity on the validation set. The mean
per-clip PSNR spread across three seeds is 2.07 dB and no single seed dominates. The
ground-truth-free medoid of Eq.~\eqref{eq:medoid} beats every individual seed on four of five
metrics and recovers most of the gap to the per-clip oracle. On the test set, best-of-3
selection adds $+0.32$ on the Stage 2
recipe (75.00 to 75.32), improving the fidelity and the distributional metrics.

\subsubsection{Prompting and Negative Prompting.}
Tab.~\ref{tab:prompt} stacks the prompt-side changes on the validation set. Structured
temporal-caption prompts outperform the plain challenge captions at either guidance setting,
and the extended negative prompt adds a further improvement on all five metrics. The test set (Tab.~\ref{tab:ladder})
shows that with the Stage 1 adapter alone, which is fine-tuned on plain captions,
structured prompting regressed from 74.76 to 74.68, and the extended negative
prompt only recovered the combination back to the level of the guidance-only configuration
(74.76). Stage 2, which aligns the training-side captions to the same format,
boosts the score to 75.00 and makes test FVD improves by 0.98, demonstrating the effectiveness of our design.


\begin{table}[t]
\begin{minipage}[t]{0.48\linewidth}
\centering
\caption{Ablation study on best-of-$N$ seed selection. The oracle selects the best sample per clip using ground truth, the medoid uses no ground truth.}
\label{tab:bestofn}
\setlength{\tabcolsep}{2pt}\scriptsize
\begin{tabular}{@{}lccccc@{}}
\toprule
Set & PSNR & SSIM & LPIPS & CLIP-S & FVD \\
\midrule
Seed 0                & 23.92 & 0.791 & 0.173 & 27.81 & 17.76 \\
Seed 1                & 23.50 & 0.780 & 0.186 & \textbf{27.94} & 21.17 \\
Seed 2                & 23.10 & 0.771 & 0.185 & 27.70 & 19.15 \\
Medoid         & \textbf{23.98} & \textbf{0.792} & \textbf{0.169} & 27.90 & \textbf{17.12} \\
\midrule
\textit{Oracle}  & \textit{24.22} & \textit{0.796} & \textit{0.166} & \textit{28.03} & \textit{17.26} \\
\bottomrule
\end{tabular}
\end{minipage}\hfill
\begin{minipage}[t]{0.48\linewidth}
\centering

\caption{Ablation study on prompting. ``struct.'' denotes the structured temporal-caption prompt and ``neg.'' the
extended negative prompt.}
\label{tab:prompt}
\setlength{\tabcolsep}{2pt}\scriptsize
\begin{tabular}{@{}lcccccc@{}}
\toprule
Prompt & $g$ & PSNR & SSIM & LPIPS & CLIP-S & FVD \\
\midrule
plain & 6           & 22.88 & 0.761 & 0.193 & 27.53 & 20.84 \\
struct. & 6        & \textbf{23.24} & \textbf{0.780} & \textbf{0.190} & \textbf{27.83} & \textbf{20.48} \\
\midrule
plain & 3            & 23.38 & 0.774 & 0.179 & 27.47 & 19.51 \\
struct. & 3          & 23.86 & \textbf{0.791} & 0.174 & 27.76 & 18.29 \\
 + neg. & 3   & \textbf{23.92} & \textbf{0.791} & \textbf{0.173} & \textbf{27.81} & \textbf{17.76} \\
\bottomrule
\end{tabular}

\end{minipage}
\end{table}

\subsubsection{Global Blending and Color Matching.}
Before adopting the motion-adaptive blend, we evaluated its global counterparts on
validation. A uniform static blend toward the last observed frame raises PSNR by a full
point (22.88 to 23.86) but adds 1.87 FVD. Matching the color statistics of
the generated frames to the history is a wash (FVD $+0.26$).
These variants suppress or recolor the moving foreground together with the background,
which is what the motion mask of Eq.~\eqref{eq:blend} avoids.

\subsubsection{Motion-Adaptive Blending.}
On validation, the blend yields the largest fidelity gains of all
post-processing variants we screened, improving PSNR, SSIM, and LPIPS simultaneously
(Tab.~\ref{tab:valablations}). On the test set, a combination with a fourth seed lifted PSNR by $0.77$, SSIM
by $0.045$, LPIPS by $0.008$, and CLIP by $0.009$ at a cost of $0.89$ FID and $1.22$ FVD,
a net gain of $+1.17$ final points compared to the previous best-of-3 version.

\subsubsection{Temporal Deflickering.}
We also evaluate the deflickering filter of Eq.~\eqref{eq:deflicker}.
On validation this improves FVD by 0.27 with all fidelity metrics held (Tab.~\ref{tab:valablations}), and on the test set
it adds a consistent $+0.035$ final points both on the single-seed Stage-2 configuration
(75.00 to 75.03) and stacked on top of best-of-3 selection (75.32 to 75.35), gaining PSNR and SSIM at a small FID cost, which is an order of
magnitude smaller than the other components.


\subsubsection{Chunked Autoregressive Rollout.}
Since the fine-tuning windows are capped at about 45 frames while test horizons reach 120
frames, we implemented a chunked autoregressive rollout that generates 32 future frames per
pass and re-conditions each subsequent pass on the last five generated frames. Despite the
training-horizon mismatch it is meant to address, the rollout scored 71.33 on the test set
against 73.31 for the otherwise identical single-pass configuration (Tab.~\ref{tab:ladder}). The damage
concentrates in FID (30.67 vs.\ 25.19) and per-frame fidelity (PSNR 18.04 vs.\ 18.60),
while FVD is roughly unchanged. Errors accumulate across chunks because later passes
condition on generated rather than observed frames, and chunk boundaries introduce visible
seams, while Cosmos 3 Nano natively supports single-pass generation of up to 200 frames at
720p. We therefore generate every clip in a single pass.

\subsubsection{Longer Conditioning History.}
The test set provides 10 to 224 history frames per clip while our recipe conditions on the
last five. Tab.~\ref{tab:valablations} extends the conditioning history at inference to 9 and 13
frames (3 and 4 latent frames), with the history
extended backward in time so that the predicted frame range is unchanged. Quality degrades
monotonically in the history length on every metric. As feeding additional latents at inference departs from its
training distribution, exploiting the longer available histories would require retraining
with a larger conditioning length.


\subsubsection{Higher LoRA Rank.}
Raising the adapter rank from 32 to 128 ($\alpha$ from 64 to 256) with an otherwise
identical Stage-1 recipe degrades four of the five validation metrics, with CLIP-S
essentially unchanged (Tab.~\ref{tab:valablations}). With 9{,}175 windows the regime is data-limited rather than capacity-limited, and
the larger adapter only overfits.

\subsubsection{Longer Stage 1 Training.}
Continuing Stage 1 from 2{,}000 to 3{,}000 and 4{,}000 iterations produces only
noise-level fidelity changes, while CLIP-S decays monotonically with training and FVD is
non-monotone across the two extra checkpoints (Tab.~\ref{tab:valablations}).
We treat the monotone CLIP-S decay as an overfitting signature, where the adapter drifts away
from prompt alignment as it memorizes the training pool.

\subsubsection{Data Mixes.}
Rebalancing the training mix to
40\% overhead views to match the test-set ratio (the unbalanced mix contains 16\%) is
slightly negative on the test set (73.05 vs.\ 73.31, Tab.~\ref{tab:ladder}). It improves the distributional
metrics (FID 24.52 vs.\ 25.19, FVD 24.36 vs.\ 24.83) but loses more on per-frame fidelity
(PSNR 18.25 vs.\ 18.60, SSIM 0.567 vs.\ 0.576).

\subsection{Qualitative Analysis}
\label{sec:qualitative}

\begin{figure}[t]
\centering
\includegraphics[width=\linewidth]{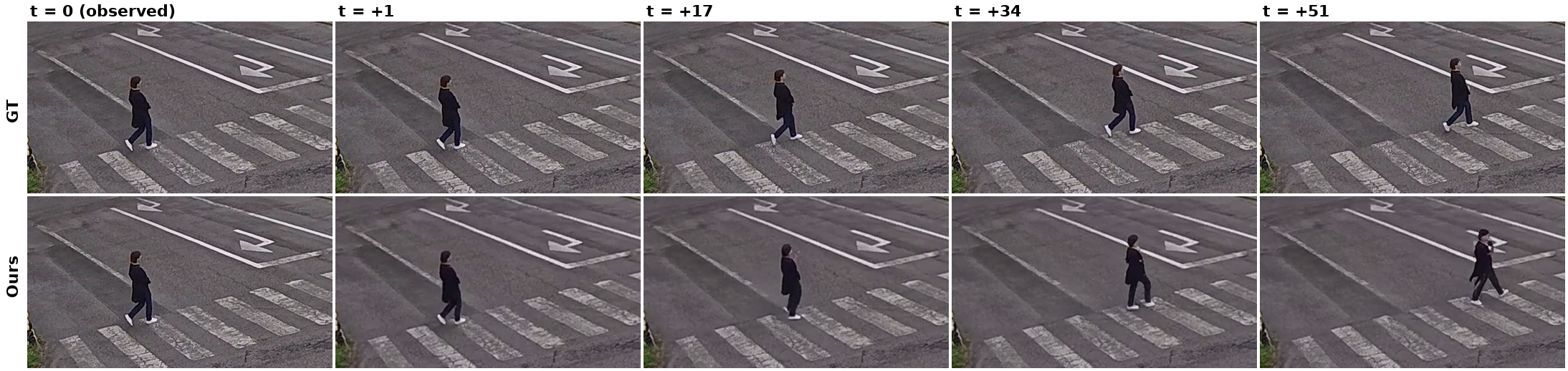}\\[2pt]
\includegraphics[width=\linewidth]{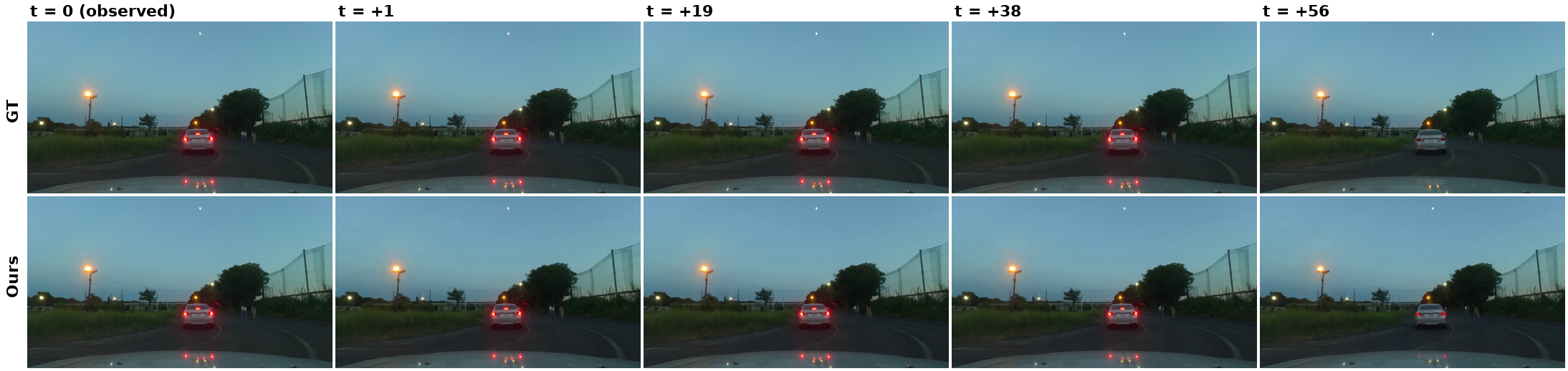}\\[2pt]
\includegraphics[width=\linewidth]{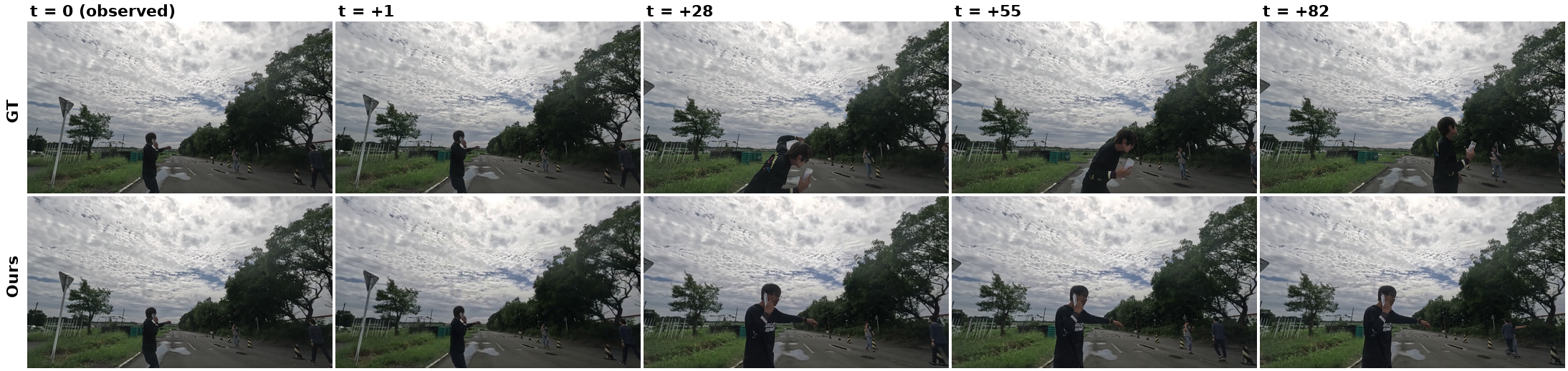}
\caption{Qualitative results of our final model on held-out clips, ground truth (top) against our prediction
(bottom). (a) Overhead view: the pedestrian is advanced along
the ground-truth trajectory across the crosswalk, with the static camera and
the road markings preserved. (b) Vehicle view at dusk: the lead vehicle, brake lights, and
light sources are continued stably. (c) Behavioral divergence: the generation stays sharp
and realistic, but the predicted pedestrian behavior differs from the ground truth.}
\label{fig:qualitative}
\end{figure}

Fig.~\ref{fig:qualitative} shows representative predictions of our final model on held-out
clips, comparing the ground-truth continuation (top rows) with our generation (bottom rows)
at four future offsets. On the overhead clip (a), a
pedestrian walks across the crosswalk over the 51-frame horizon. Our prediction advances
him along the ground-truth trajectory, keeping his position close to the reference at every
offset while the camera stays perfectly static and the road markings remain sharp, although
his gait and arm pose drift out of phase by the end of the horizon. On the dusk vehicle clip (b), the model continues the approach toward the lead
car with its brake lights and the surrounding light sources rendered stably over time; the
street lamps neither bloom nor drift in color, the failure modes explicitly targeted by our
extended negative prompt. Clip (c) illustrates the main remaining error source: behavioral
divergence. The generation is sharp and physically plausible throughout, but the model
predicts that the nearby pedestrian keeps standing while in the ground truth he leans down
toward the camera, so the per-frame fidelity metrics penalize the clip heavily even though
the video itself is realistic. Divergence of this kind is exactly the stochasticity that our
medoid selection exploits: futures on which independent samples agree are much less likely
to contain such idiosyncratic behavior errors.

\begin{figure}[t]
\centering
\includegraphics[width=\linewidth]{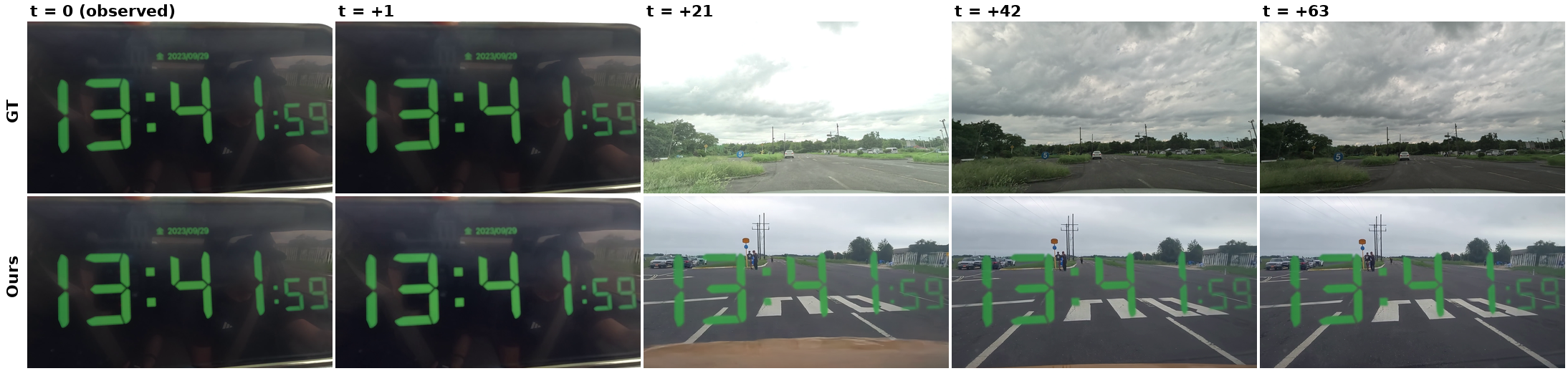}
\caption{Visual conditioning dominates the prompt. The history frames of this validation
clip show a dashboard clock; the ground truth (top) cuts to the road scene, whereas our
prediction (bottom) carries the glowing digits into the generated road scene and holds
them for the whole horizon. The prompt describes only the road scene.}
\label{fig:clock}
\end{figure}

Beyond these cases, we observe two systematic behaviors. First, prediction quality is
consistently higher on overhead clips than on vehicle clips, since ego-motion makes the
whole frame non-static and leaves no anchor for the background, which also makes the
motion-adaptive blend self-deactivate on vehicle clips as intended. Second, when the visual
history conflicts with the textual description, the visual conditioning dominates. Fig.~\ref{fig:clock} shows the clearest instance: the clip opens on a close-up of a
dashboard clock before cutting to the road, and although the prompt describes only the road
scene, the model carries the glowing digits into the generated continuation and keeps them
superimposed for the entire horizon, while the ground truth cuts away cleanly. This behavior indicates that the five conditioning frames, not the prompt,
anchor the generated scene.

\subsection{Performance in the Challenge}
\label{sec:challenge}

Tab.~\ref{tab:ladder} reports the progression of our submissions on the official test set. Guidance
tuning, Stage 2 caption alignment, medoid selection, and the final blending step each
improve the final score, and the last step is the largest ($+1.17$).

Tab.~\ref{tab:leaderboard} shows the final public leaderboard of the AI City Challenge 2026 Track~5. Our solution (Qyn) ranks first
with 76.49, 0.45 points ahead of the second-ranked team. The metric profile reflects the design of
our pipeline: we attain the best PSNR, LPIPS, CLIP, and (up to $10^{-4}$) SSIM of all teams,
while the second-ranked team leads the distributional metrics FID and FVD.

\begin{table}[t]
\centering
\caption{Final AI City Challenge Track~5 public leaderboard (top 5 only).}
\label{tab:leaderboard}
\setlength{\tabcolsep}{4pt}
\begin{tabular}{@{}clccccccc@{}}
\toprule
Rank & Team & Final$\uparrow$ & PSNR$\uparrow$ & SSIM$\uparrow$ & LPIPS$\downarrow$ & CLIP$\uparrow$ & FID$\downarrow$ & FVD$\downarrow$ \\
\midrule
\textbf{1} & \textbf{Qyn (ours)} & \textbf{76.49} & \textbf{20.12} & \textbf{0.650} & \textbf{0.246} & \textbf{0.950} & 22.41 & 21.79 \\
2 & SSUPER              & 76.04 & 19.73 & 0.630 & 0.249 & 0.938 & \textbf{21.16} & \textbf{19.46} \\
3 & Latent Painter      & 75.13 & 19.72 & 0.650 & 0.266 & 0.945 & 26.52 & 24.79 \\
4 & CHTTL\_A30          & 74.05 & 18.86 & 0.597 & 0.281 & 0.942 & 23.78 & 24.11 \\
5 & VGU\_ai\_lab        & 73.28 & 19.74 & 0.647 & 0.294 & 0.945 & 33.61 & 29.35 \\
\bottomrule
\end{tabular}
\end{table}

\section{Conclusion}

In this paper, we presented CosmosAlign, the first-place solution to Track~5 of the AI
City Challenge 2026. Starting from the pretrained Cosmos3-Nano world model, we adapt it
with a two-stage LoRA recipe that aligns the conditioning-mode distribution and the
caption format with the evaluated task, complemented by a training-free test-time
procedure of consensus-based medoid selection and motion-adaptive blending. Future works can strengthen the text-conditioning pathway, for example through
explicit behavior controls or caption-conditioned selection among sampled futures, since
the remaining errors stem from generated agent behavior diverging from the description
and from the visual conditioning dominating the prompt. Another direction is to retrain with
longer conditioning lengths to exploit the full provided history, which our ablations
show cannot be used by an adapter trained on five frames, and to verify that our
alignment-oriented recipe generalizes beyond a single backbone by applying it to larger
variants such as Cosmos3-Super, and other world model families.

\bibliographystyle{splncs04}
\bibliography{main}
\end{document}